\PassOptionsToPackage{numbers, square}{natbib}
\documentclass{article} 
\usepackage[preprint]{neurips_2020}
\usepackage[utf8]{inputenc}
\usepackage{amsmath}
\newcommand\numberthis{\addtocounter{equation}{1}\tag{\theequation}}
\usepackage{amssymb}
\usepackage{amsfonts}
\usepackage{graphicx}
\usepackage{titlesec}

\usepackage{mathtools}
\DeclarePairedDelimiterX{\infdivx}[2]{\big(}{\big)}{%
  #1\;\delimsize\|\;#2%
}

\usepackage{xcolor}
\usepackage{hyperref}
\hypersetup{colorlinks=true,
linkcolor = blue,
linkbordercolor = {white},
citecolor = blue}

\usepackage{url}

\title{Complex Skill Acquisition Through Simple Skill Imitation Learning}

\author{Pranay Pasula \\
Department of Electrical Engineering and Computer Sciences\\
University of California, Berkeley\\
\texttt{pasula@berkeley.edu}
}

\begin{document}

\maketitle

%

\author{%
  Pranay Pasula \\
  Department of Electrical Engineering and Computer Science\\
  University of California, Berkeley\\
  \texttt{pasula@berkeley.edu} \\
}

\begin{abstract}
  Humans often think of complex tasks as combinations of simpler subtasks in order to learn those complex tasks more efficiently. For example, a backflip could be considered a combination of four subskills: jumping, tucking knees, rolling backwards, and thrusting arms downwards. Motivated by this line of reasoning, we propose a new algorithm that trains neural network policies on simple, easy-to-learn skills in order to cultivate latent spaces that accelerate imitation learning of complex, hard-to-learn skills. We focus on the case in which the complex task comprises a \emph{concurrent} (and possibly \emph{sequential}) combination of the simpler subtasks, and therefore our algorithm can be seen as a novel approach to \emph{concurrent hierarchical imitation learning}. We evaluate our algorithm on difficult tasks in a high-dimensional environment and find that it consistently outperforms a state-of-the-art baseline in training speed and overall performance.
\end{abstract}

\section{Introduction}

Humans have the power to reason about complex tasks as combinations of simpler, interpretable subtasks. There are many hierarchical reinforcement learning approaches designed to handle tasks comprised of sequential subtasks \cite{sutton1999between, konidaris2007building}, but what if a task is made up of \textit{concurrent} subtasks? For example, someone who wants to learn to do a backflip may consider it to be combination of sequential \textit{and} concurrent subtasks: jumping, tucking knees, rolling backwards, and thrusting arms downwards. Little focus has been given to designing algorithms that decompose complex tasks into distinct concurrent subtasks. Even less effort has been put into finding decompositions that are made up of independent yet interpretable concurrent subtasks, even though analogous approaches have been effective on many challenging artificial intelligence problems \cite{chen2016infogan, burgess2018understanding}.

We believe that endowing intelligent agents with the ability to disentangle complex tasks into simpler, distinct, and interpretable subtasks could help them learn complex tasks more efficiently. Furthermore, by explicitly encouraging agents to learn distinct, interpretable subtasks, we would incline the agents towards learning distinct and interpretable representations, which can be leveraged in powerful ways. For example, in the context of algorithmic human-robot interaction, an agent, embodied or virtual, may be able to use these representations to learn complex tasks a human cannot perform well but are comprised of simpler concurrent subtasks at which the human is proficient (e.g., backflip decomposition). Another example is that the agent could continuously interpolate between subtasks to adjust its behavior in semantically-meaningful ways \cite{chen2016infogan, burgess2018understanding}. This is particularly motivating because such fine-scale interpolations are characteristic of humans.

We propose a new generative model for encoding and generating arbitrarily complex trajectories. We augment the VAE objective used in \cite{wang2017robust} in order to induce latent space structure that captures the relationship between a behavior and the subskills that comprise this behavior in a disentangled and interpretable way. Furthermore, we derive an approximation to this augmentation that circumvents the need for potentially expensive variational inference. We evaluate both the original and modified objectives on challenging imitation learning problems, in which agents are trained to perform complex behaviors after being trained on subskills that qualitatively comprise those behaviors.

\section{Embedding and reconstructing trajectories}

We use a conditional variational autoencoder (CVAE) \cite{sohn2015learning, kingma2013auto} to learn a semantically-meaningful low-dimensional embedding space that can (1) help an agent learn new behaviors more quickly, (2) be sampled from to generate behaviors, (3) and shed light on high-level factors of variation (e.g. subskills) that comprise complex behaviors.

\begin{figure}[h]
    \includegraphics[scale=0.8]{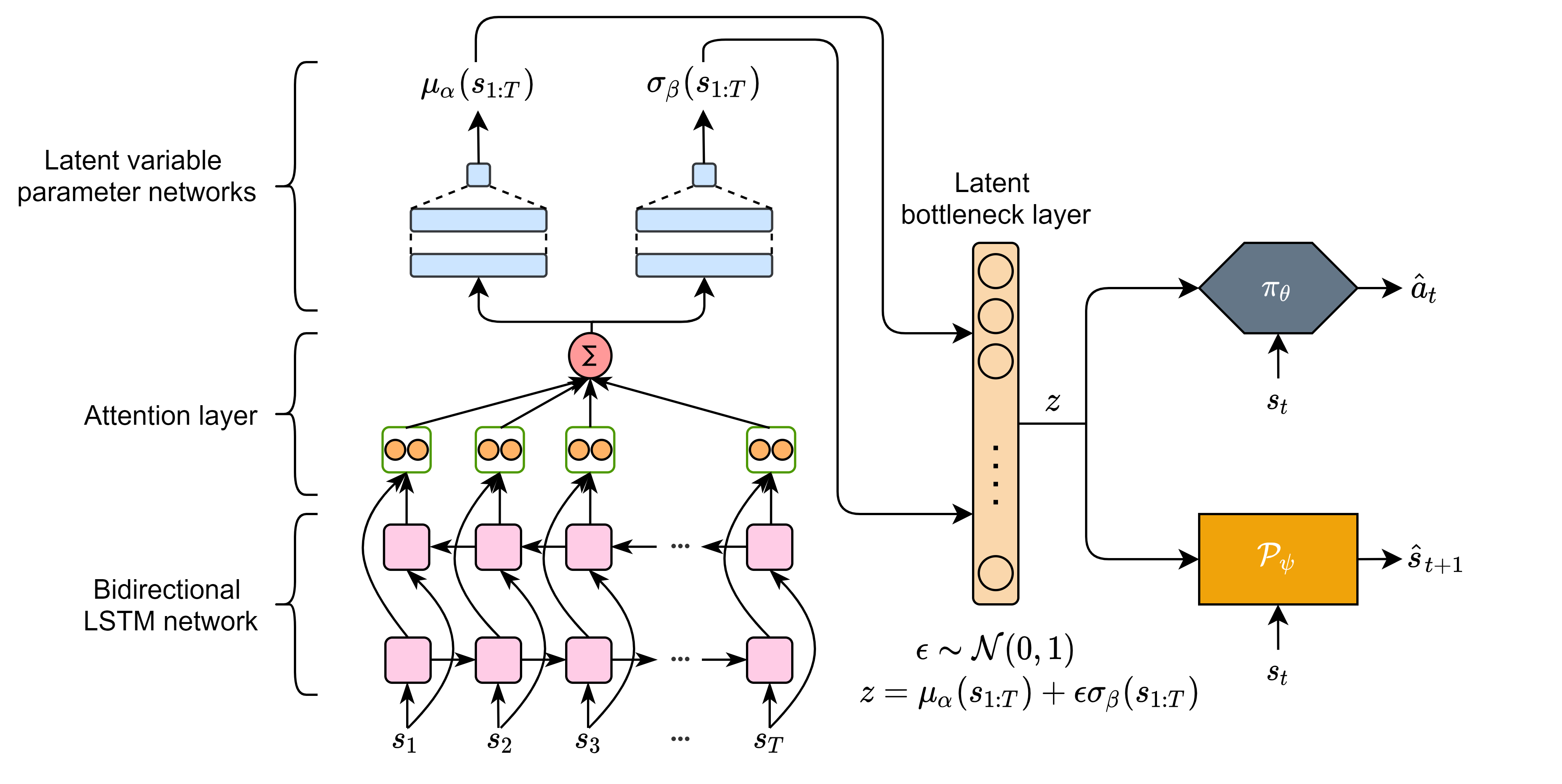}
    \centering
    \caption{The conditional VAE we use to encode and generate trajectories. The \emph{bidirectional LSTM network}, \emph{attention layer}, and \emph{latent variable parameter networks} comprise the \emph{encoder} $q_\phi(z | s_{1:T}$), which for notational convenience we use interchangeably with $q_\phi(z | \tau).$ To generate state-action sequences, or trajectories, we sample noise $\epsilon \thicksim \mathcal{N}(0,1)$ and compute latent vector $z$. Then we condition \emph{policy} $\pi_\theta$ and \emph{dynamics model} $\mathcal{P}_\psi$ on $z$ and $s_t$ for each timestep $t=1,2,3,\dotsc,T-1$ to output $\tilde{\tau} = (\hat{s_1}, \hat{a_1}, \hat{s_2}, \hat{a_2}, \dotsc, \hat{s_T}, \hat{a_T}).$}
    \label{fig:cvae}
\end{figure}

Illustrated by Figure \ref{fig:cvae}, our CVAE has a bi-directional LSTM (BiLSTM) \cite{hochreiter1997long, schuster1997bidirectional} state-sequence encoder $q_\phi(z|s_{1:T})$, an attention module \cite{bahdanau2014neural, zhou2016attention} that maps the BiLSTM output to values that parametrize the distribution from which the latent (i.e. trajectory) embedding $z$ is sampled, a conditional WaveNet \cite{oord2016wavenet} state decoder $\mathcal{P}_\psi(s_{t+1}|s_t, z)$, which serves as a \emph{dynamics model}, and a multi-layer perceptron (MLP) action decoder $\pi_\theta(a_t|s_t, z)$, which serves as a \emph{policy} whose outputs parametrize the normal distribution from which $a_t$ is sampled. The bidirectional LSTM network captures sequential information over the states of the trajectories, and the conditional WaveNet allows for exact density modeling of the possibly multi-modal dynamics.

We can train this CVAE by minimizing the following objective
\vspace{-1.0ex}
\begin{multline}\label{orig_vae_objec}
    \mathcal{L}(\theta, \phi, \psi; \tau^i) = - \mathbb{E}_{z \sim q_\phi(z|s_{1:T_i}^i)} \left[ \sum_{t=1}^{T_i} \log \pi_\theta(a_t^i|s_t^i,z) + \log \mathcal{P}_\psi(s_{t+1}^i | s_t^i, z) \right] \\
    + D_{KL}\infdivx{q_\phi(z|s_{1:{T_i}}^i)}{p(z)}.
\end{multline}
In Section 3 we will modify this objective in order to encourage the latent space to capture semantically meaningful relationships between complex behaviors and the subskills that comprise those behaviors.

\section{Shaping the latent (i.e. trajectory embedding) space}

Some skills can be seen as approximate combinations of certain subskills. Training a VAE to embed and reconstruct demonstrations of these skills and subskills using (\ref{orig_vae_objec}) would generally result in an embedding space with no clear relationship between skill and subskill embedding, especially if the dimensionality of the latent space is large or the number of demonstrated behaviors is small.

\begin{figure}[h]
    \includegraphics[scale=0.6]{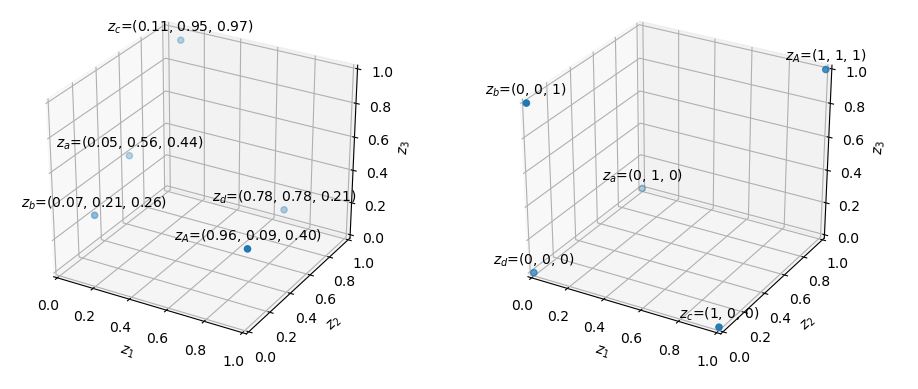}
    \centering
    \caption{An example of latent space restructuring. \textit{Left:} original latent space. \textit{Right:} hypothetical latent space induced by our approach (created intentionally for illustrative purposes).}
    \label{fig:restruc}
\end{figure}

Motivated by semantically meaningful latent representations found in other work \cite{mikolov2013distributed}, we aim to induce a latent space structure so that a behavior embedding is the sum of the its subskill embeddings. Concretely, if $z_A$ is a backflip embedding and $z_a, z_b, z_c, z_d$ are embeddings corresponding to jumping, tucking knees, rolling backwards, and thrusting arms downwards, we want to have $z_A = z_a + z_b + z_c + z_d$. An example of such latent space restructuring is shown in Figure \ref{fig:restruc}.
However, the VAE models probability distributions, so enforcing equality between one instance of a behavior and one instance of its subskills is insufficient. Instead, we want the random variables (RVs) representing the embeddings of the subskills to relate to the RV representing the embedding of the behavior comprised of those subskills. Another way to do this is to relate the subskill embedding RVs with the RV representing the trajectory generated by decoder networks $\mathcal{P}_\psi$ and $\pi_\theta$ when conditioned on an embedding of the corresponding complex behavior, as shown in \ref{fig:cvae}.

Suppose $\tau_V$ is a behavior comprised of $M$ subskills $\{\tau_{(1)}, \tau_{(2)}, \dotsc, \tau_{(M)}\}.$ Let $\tilde{\tau}_{v}=(\hat{s}_1, \hat{a}_1, \hat{s}_2, \hat{a}_2, \dotsc, \hat{s}_T, \hat{a}_T)_{v}$ represent the trajectory generated from an embedding corresponding to $\tau_{v}$. Define $V = z_1 + z_2 + \cdots + z_M$, where $z_i \sim q_\phi(z|s_{(i), \; 1:T_{(i)}})$. To train the encoder $q_\phi(z|s_{1:T})$, state decoder $\mathcal{P}_\psi(\hat{s}_{t+1}|s_{t-1},z)$, and action decoder $\pi_\theta(\hat{a}_t|s_t, z)$ simultaneously, we aim to maximize the mutual information between $V$ and $\tilde{\tau}_{v}$, which can be expressed as
\begin{align*}
    I(V; \tilde{\tau}_v) & = H(V) - H(V | \tilde{\tau}_v)
    \numberthis \label{eq:mi1}\\[0.5ex]
    & = -\mathbb{E}_{V \sim p(V)} \big[ \log p(V) \big] + \mathbb{E}_{V \sim p(V | \tilde{\tau}_v)} \big[ \log p(V | \tilde{\tau}_v) \big]. 
    \numberthis \label{eq:mi2}
\end{align*}
If the latent variable prior distribution $p(z_i)$ is Gaussian for all $i=1,2,3,...,M$, the entropy $H(V)$ is easy to compute, with an analytical solution under minor assumptions. We describe how to evaluate $H(V)$ in \ref{s:ent}. Going forward we omit the subscript in $\tilde{\tau}_v$ and hats in $\mathcal{P}(\hat{s}_{t+1}|s_t, z)$ and $\pi(\hat{a}_t|s_t,z)$ unless they are needed to distinguish different quantities. 

\subsection{Lower bounding mutual information through variational inference}

We can't compute (\ref{eq:mi2}) directly because we don't have access to the true posterior distribution $p(V|\tilde{\tau})$. Therefore, we instead introduce a distribution $Q(V | \tilde{\tau})$ as a variational approximation to $p(V | \tilde{\tau})$ to get $L_I(\tilde{\tau}, Q)$, a variational lower bound of $I(V; \tilde{\tau}),$ 
\vspace{-2.0ex}
\begin{align*}
    L_I(\tilde{\tau}, Q) & = \mathbb{E}_{V \sim p(V), \tau \sim \tilde{\tau} | V} \big[ \log Q(V | \tau) \big] + H(V) \\[1.0ex]
    & = \mathbb{E}_{\tau \sim \tilde{\tau}} \big[ \mathbb{E}_{V \sim p(V | \tau)} \big[\log Q(V | \tau) \big] \big] + H(V) \\[1.0ex]
    & \leq I(V; \tilde{\tau})
\end{align*}

in an approach similar to that of \cite{chen2016infogan}.

But, unlike in \cite{chen2016infogan}, $Q(V|\tilde{\tau})$ is \emph{not} the same as $q(z|s_{1:T})$, the distribution approximated by the encoder network $q_\phi$ in our CVAE. Furthermore, even though embedding variables $z_1, z_2, ..., z_M$ are independent, they are \emph{not} conditionally independent given $\tilde{\tau}$. Therefore we \emph{cannot} simply replace $Q(V|\tilde{\tau})$ with $\sum_{i=1}^{M} q(z_i | \tilde{\tau})$ and instead may need to again use variational inference to find $Q(V | \tilde{\tau}),$ which requires training an additional VAE. Fortunately by learning a reasonably good approximation to $p(z|\tilde{\tau})$, we can avoid this additional expense.

We now derive a result that allows us to bypass variational inference.

\subsection{Lower bounding mutual information without variational inference}

We derive a simpler lower bound to $I(V;\tilde{\tau})$ that allows us to circumvent the time and memory costs associated with training a VAE to model $Q(V|\tilde{\tau})$.

For clarity in the following derivation, let $V_p = \sum_{i=p}^{M} z_i.$ Then we have
\begin{align*}
    H(V | \tilde{\tau}) &= H(V_1 | \tilde{\tau}) \\[0.5ex]
    & = H(z_1 + z_2 + \cdots + z_M | \tilde{\tau}) \\[0.5ex]
    & = H(z_1 | \tilde{\tau}) + H(z_1 + z_2 + \cdots + z_M | z_1, \tilde{\tau}) - H(z_1 | z_1 + z_2 + \cdots + z_M, \tilde{\tau}) \\[0.5ex]
    & = H(z_1 | \tilde{\tau}) + H(z_2 + z_3 + \cdots + z_M | z_1, \tilde{\tau}) - H(z_1 | z_1 + z_2 + \cdots + z_M, \tilde{\tau}) \\[0.5ex]
    & \leq H(z_1 | \tilde{\tau}) + H(z_2 + z_3 + \cdots + z_M | \tilde{\tau}) - H(z_1 | z_1 + z_2 + \cdots + z_M, \tilde{\tau}) \\[0.5ex]
    & = H(z_1 | \tilde{\tau}) + H(V_2 | \tilde{\tau}) - H(z_1 | V_1, \tilde{\tau}) 
\end{align*}

By rolling out $H(V_p | \tilde{\tau})$ recursively for $p = 1, 2, 3,..., M-1$, we get
\begin{align*}
    H(V | \tilde{\tau}) & \leq \sum_{i=1}^{M} \left[ H(z_i | \tilde{\tau}) - H(z_i | V_i, \tilde{\tau}) \right] \\
    & \leq \sum_{i=1}^{M} H(z_i | \tilde{\tau}) \\
    & = \sum_{i=1}^{M} -\mathbb{E}_{z_i \sim p(z_i | \tilde{\tau})} \left[ \log p( z_i | \tilde{\tau}) \right] \\
    & \approx \sum_{i=1}^{M} -\mathbb{E}_{z_i \sim q_{\phi}(z_i | \tilde{\tau})} \left[ \log q_{\phi}( z_i | \tilde{\tau}) \right]
\end{align*}
if $p(z | \tilde{\tau}) \approx q_{\phi}(z | \tilde{\tau}).$ Plugging this result into (\ref{eq:mi1}) allows us to lower bound $I(V; \tilde{\tau})$ as follows,
\begin{align*}
    I(V;\tilde{\tau}) & \geq -\mathbb{E}_{V \sim p(V)} \big[ \log p(V) \big] + \sum_{i=1}^{M} \mathbb{E}_{z_i \sim p(z_i | \tilde{\tau})} \left[ \log p( z_i | \tilde{\tau}) \right] \\
    & \approx -\mathbb{E}_{V \sim p(V)} \big[ \log p(V) \big] + \sum_{i=1}^{M} \mathbb{E}_{z_i \sim q_{\phi}(z_i | \tilde{\tau})} \left[ \log q_{\phi}( z_i | \tilde{\tau}) \right],
\intertext{and we can obtain an unbiased estimate of the second term by sampling $z_i \sim q_\phi(z_i | \tilde{\tau}) $ to get}
    I(V; \tilde{\tau}) & \ \gtrapprox \ -\mathbb{E}_{V \sim p(V)} \big[ \log p(V) \big] + \frac{1}{N} \sum_{n=1}^{N} \sum_{i=1}^{M} \log q_{\phi}(z_{n,i} | \tilde{\tau}),
    \numberthis \label{eq:sample_est}
\end{align*}
in which $x \gtrapprox y$ denotes that $x$ is approximately greater than or equal to $y.$

By maximizing the lower bound in (\ref{eq:sample_est}), we (approximately) maximize $I(V; \tilde{\tau})$.

\subsubsection{Entropy evaluation}
\label{s:ent}
Computing the entropy for an arbitrary distribution may be difficult, but by setting $X$ to be a Gaussian RV---the standard choice for VAE encoders---$H(X)$ has the simple, closed-form expression
\begin{align*}
    H(X) = \frac{1}{2}(1 + \ln (2 \pi \sigma_{X}^2)),
\end{align*}

where $\sigma_X$ is the standard deviation of $X$. We choose $q_\phi(z|s_{1:T})$ to parametrize a Gaussian distribution and assume that state sequences from different subskills are sufficiently unrelated so that they can be considered statistically independent. This is generally a safe assumption because even minor differences in subskills will tend to place trajectories corresponding to different skills in very different locations within the trajectory space. It follows that $V$ is the sum of Gaussian RVs and has the simple form
\begin{align*}
    V \sim \mathcal{N}(\mu_{z_a} + \mu_{z_b} + \cdots + \mu_{z_M}, \; \sigma_{z_a}^2 + \sigma_{z_b}^2 + \cdots + \sigma_{z_M}^2),
\end{align*}

and the entropy of $V$ is
\begin{align*}
    H(V) = \frac{1}{2}(1 + \ln(2 \pi (\sigma_{z_a}^2 + \sigma_{z_b}^2 + \cdots + \sigma_{z_M}^2))).
    \numberthis \label{eq:sum_gauss_ent}
\end{align*}
\subsection{Regularization with variational approximation}
To encourage a semantically meaningful relationship between a behavior embedding and this behavior's subskill embeddings, we regularize the objective  in (\ref{orig_vae_objec}) with $L_I(\tilde{\tau}, Q_\alpha)$ to get
\begin{multline}\label{reg_vae_objec}
    \mathcal{L}(\theta, \phi, \psi; \tau^i) = - \mathbb{E}_{z \sim q_\phi(z|s_{1:T_i}^i)} \left[ \sum_{t=1}^{T_i} \log \pi_\theta(a_t^i|s_t^i,z) + \log \mathcal{P}_\psi(s_{t+1}^i | s_t^i, z) \right] \\
    + D_{KL}\infdivx{q_\phi(z|s_{1:{T_i}}^i)}{p(z)} + \lambda L_I(\tilde{\tau}, Q_\alpha),
\end{multline}

where $\lambda > 0$ is a hyperparameter that controls the trade-off between original objective and degree of latent space shaping.

\subsection{Regularization without variational approximation}
If we want to avoid performing potentially expensive variational inference, we can use (\ref{eq:sample_est}), the result we derived earlier, in place of $L_I(\tilde{\tau}, Q)$,
\begin{multline}\label{reg_vae_objec_2}
    \mathcal{L}(\theta, \phi, \psi; \tau^i) = - \mathbb{E}_{z \sim q_\phi(z|s_{1:T_i}^i)} \left[ \sum_{t=1}^{T_i} \log \pi_\theta(a_t^i|s_t^i,z) + \log \mathcal{P}_\psi(s_{t+1}^i | s_t^i, z) \right] \\
    + D_{KL}\infdivx{q_\phi(z|s_{1:{T_i}}^i)}{p(z)} + \lambda \left( -\mathbb{E}_{V \sim p(V)} \big[ \log p(V) \big] + \frac{1}{N} \sum_{n=1}^{N} \sum_{i=1}^{M} \log q_{\phi}(z_{n,i} | \tilde{\tau}) \right).
\end{multline}

As shown in \ref{s:ent}, the inner expectation in (\ref{reg_vae_objec_2}) can be evaluated analytically if the latent variables $\{z_i\}_{i=1}^{M}$ are independent and normally distributed---the standard case with VAEs.

\section{Experiments and results}

\begin{figure}[h]
    \includegraphics[scale=0.45]{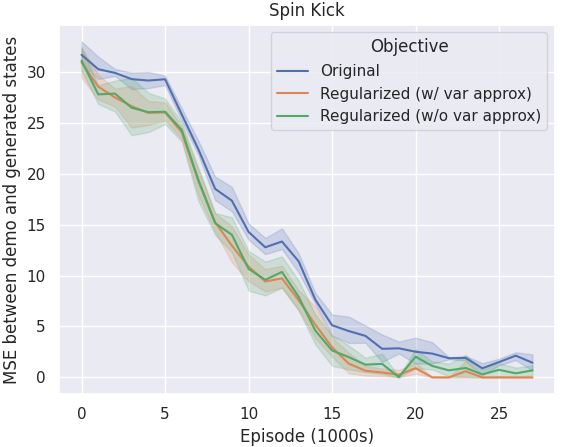} \includegraphics[scale=0.41]{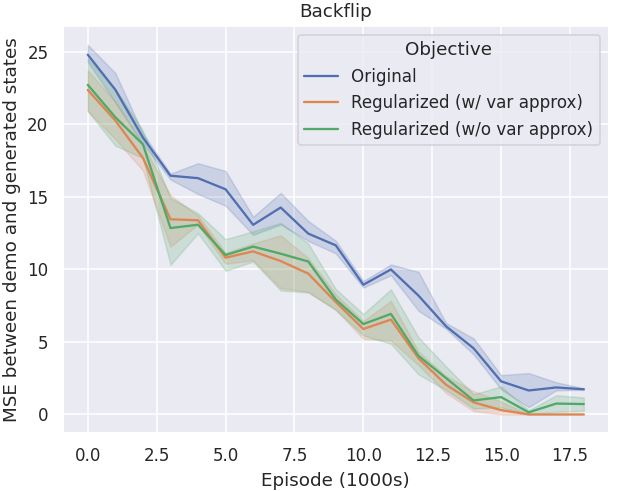}
    \centering
    \caption{MSE (lower is better) between demonstration states and generated states on the DeepMimic \emph{spin kick} and \emph{backflip} tasks averaged over 5 different random seeds. \textbf{\emph{Regularized} denotes \emph{our} approaches (\ref{reg_vae_objec}), (\ref{reg_vae_objec_2}), and \emph{Original} denotes the \emph{state-of-the-art} baseline (\ref{orig_vae_objec}).}}
    \label{fig:results}
\end{figure}

We evaluate our approach on a 197-dimensional state and 34-dimensional action space humanoid simulated in Bullet \cite{coumans2015bullet}. We use policies that were pre-trained by \cite{peng2018deepmimic} to perform \emph{kick}, \emph{spin}, and \emph{jump}, as subskills that qualitatively comprise the behavior \emph{spin kick}. We also take a similar approach for the behavior \emph{backflip}. We train three sets of five VAEs on the subskills: one set optimizes for the original VAE objective (\ref{orig_vae_objec}), another set optimizes for the objective regularized by the variational approximation (\ref{reg_vae_objec}), and the third set optimizes for the objective regularized without variational inference. To compare the proposed approach with the original, we evaluate the training process of each set of VAEs by considering the similarity between the generated trajectories and the pre-trained \emph{spin kick} and \emph{backflip} policy demonstrations. Results of the mean squared error (MSE) between the generated and demonstration states averaged over 5 different random seeds are shown in Figure \ref{fig:results}.
We find that our approaches attain better overall performance and train faster than the baseline algorithm. This suggests that we can bootstrap the learning of difficult tasks by training agents on simpler, related subtasks while inclining their representations toward certain hierarchical structures.

\section{Discussion and future work}

We explored the idea of inducing certain latent structure through the maximization of mutual information between generated behaviors and embeddings of the subskills that qualitatively comprise those behaviors, which, to the best of our knowledge, has not yet been investigated. Though our algorithm outperformed the state-of-the-art baseline, there is much room for future work. The CVAE could be replaced with a $\beta$-CVAE \cite{higgins2017beta} to control disentanglement of $z$. The proposed approach could be evaluated on behaviors and subskills that more strictly adhere to concurrent relationship desired. A larger number of behaviors, such as those put forth by \cite{yu2020meta}, could be trained at once, both to constrain the latent space and to enrich the pool of subskills from which to train on and inspect relationships between. The non-variational mutual information approximation could be compared to the variational one in order to quantify accuracy. Interpolations within the convex hull of subskill embeddings could be used to fine-tune known behaviors or generate completely new behaviors.

\bibliography{neurips_2020}

\end{document}